\documentclass{article}

\usepackage[preprint]{neurips_2026}
\usepackage[utf8]{inputenc}
\usepackage[T1]{fontenc}
\usepackage{url}
\usepackage{xspace}
\usepackage{microtype}
\usepackage{amsmath,amssymb,amsthm}
\usepackage{graphicx}
\usepackage{booktabs}
\usepackage{multirow}
\usepackage{array}
\usepackage[table]{xcolor}
\usepackage{caption}
\usepackage{subcaption}
\usepackage{algorithm}
\usepackage{algpseudocode}

\title{\vspace{-0.5em}Is Complex Training Necessary for Long-Tailed OOD Detection? A Re-think from Feature Geometry}
\author{%
  Ningkang Peng \\
  Nanjing Normal University
  \And
  Xuanming Chen \\
  Nanjing Normal University
  \And
  Yanhui Gu \\
  Nanjing Normal University
}
\date{}

\newcommand{\ID}{\mathrm{ID}}
\newcommand{\OOD}{\mathrm{OOD}}
\newcommand{\E}{\mathbb{E}}
\newcommand{\R}{\mathbb{R}}
\newcommand{\Energy}{\mathcal{E}}
\newcommand{\HPM}{\ifmmode\mathrm{HPM}\else\(\mathrm{HPM}\)\xspace\fi}
\newcommand{\PMD}{\ifmmode\mathrm{RP}\text{-}\mathrm{MD}\else\(\mathrm{RP}\text{-}\mathrm{MD}\)\xspace\fi}
\newcommand{\SMD}{\ifmmode\mathrm{HC}\text{-}\mathrm{MD}\else\(\mathrm{HC}\text{-}\mathrm{MD}\)\xspace\fi}
\newcommand{\VMD}{\ifmmode\mathrm{Mahalanobis}\else Mahalanobis\xspace\fi}
\newcommand{\row}{\mathrm{row}}
\newcommand{\nullsp}{\mathrm{null}}
\definecolor{bestgreen}{RGB}{84,148,96}
\definecolor{secondgreen}{RGB}{154,205,160}
\definecolor{thirdgreen}{RGB}{218,238,218}
\newcommand{\bestcell}[1]{\cellcolor{bestgreen!85}\textbf{#1}}
\newcommand{\secondcell}[1]{\cellcolor{secondgreen!85}\underline{#1}}
\newcommand{\thirdcell}[1]{\cellcolor{thirdgreen}#1}

\theoremstyle{definition}

\newtheorem{lemma}{Lemma}
\newtheorem{theorem}{Theorem}

\begin{document}
\maketitle

\begin{abstract}
Long-tailed out-of-distribution (LT-OOD) detection is often addressed with specialized training, including auxiliary out-of-distribution (OOD) data, abstention heads, contrastive objectives, energy losses, or gradient-conflict control.
We show that these training mechanisms can obscure a simpler issue: frozen long-tailed representations may already contain useful OOD evidence, but raw Mahalanobis distance is distorted by frequency-coupled feature radius and poorly supported tail covariance.
We propose \emph{Hyperspherical Pooled Mahalanobis} (\HPM), a post-hoc detector that normalizes features onto the unit sphere and replaces class-specific covariance with a pooled, ridge-regularized metric while keeping class means as semantic anchors.
In CIFAR-LT experiments and an ImageNet-100-LT near-OOD boundary analysis, \HPM improves raw Mahalanobis scoring; for Prior-Calibrated ERM (PC-ERM), it raises AUROC from 46.49 to 85.67 on CIFAR-10-LT and from 50.40 to 78.35 on CIFAR-100-LT.
This simple PC-ERM+\HPM pipeline also achieves the best Log Efficiency Score (LES; 3.08) on CIFAR-100-LT, retaining roughly 95\% of the best CIFAR-100-LT AUROC observed among the compared post-hoc scores at substantially lower training-time cost.
These results argue for evaluating representation quality, detector geometry, and training complexity as separate factors in LT-OOD detection.
\end{abstract}

\section{Introduction}

Open-world recognition must do more than name the classes seen during training. A deployed system also needs to recognize when an input falls outside the in-distribution (ID) training distribution~\citep{hendrycks2017baseline,liang2018odin,lee2018mahalanobis}. This requirement is especially sharp in long-tailed domains such as autonomous perception, medical triage, ecological monitoring, and industrial inspection, where rare ID categories and novel hazards coexist. Long-tailed out-of-distribution (LT-OOD) detection therefore asks for two behaviors at once: retain rare known classes and reject inputs from outside the training distribution.

Because head classes provide dense supervision while tail classes may have only a few examples~\citep{liu2019imagenetlt,cao2019ldam}, many recent LT-OOD methods respond by making training more specialized. Some use energy-based objectives~\citep{liu2020energy} or auxiliary outlier exposure~\citep{hendrycks2019oe}; others introduce contrastive tail-OOD separation~\citep{wang2022pascl}, abstention or calibrated outlier classes~\citep{wei2024eat,miao2024cocl}, adaptive outlier distributions~\citep{miao2024adaptod}, gradient-conflict mitigation~\citep{zhang2025darl}, or semantic tail prioritization~\citep{he2025patt}. These methods have advanced the benchmark, but they raise a basic question: is increasingly complex training necessary for strong LT-OOD detection, or can a simple long-tail classifier already be competitive when the detector uses its representation properly?

We study this question through Prior-Calibrated ERM (PC-ERM), a simple baseline trained with cross-entropy and long-tail prior calibration~\citep{menon2021longtail}. PC-ERM uses no auxiliary OOD data, abstention head, contrastive objective, or specialized rejection loss. To compare accuracy with cost, we also report a Log Efficiency Score (LES), which relates the strongest post-hoc OOD AUROC of a trained model to its training time. LES separates absolute OOD performance from compute-normalized performance.

Our experiments point to a different bottleneck: raw Mahalanobis detection reads long-tailed features through an unstable geometry. Feature radius can be coupled with class frequency, so raw distance mixes semantic displacement with a training-frequency artifact. Tail classes also have too few samples to support class-specific covariance, making the estimated precision low-rank, ill-conditioned, and sensitive to regularization. Thus a simple classifier may look weak under raw distance not because its representation is unusable, but because the post-hoc metric is distorted.

\begin{figure}[t]
    \centering
    \includegraphics[width=\linewidth]{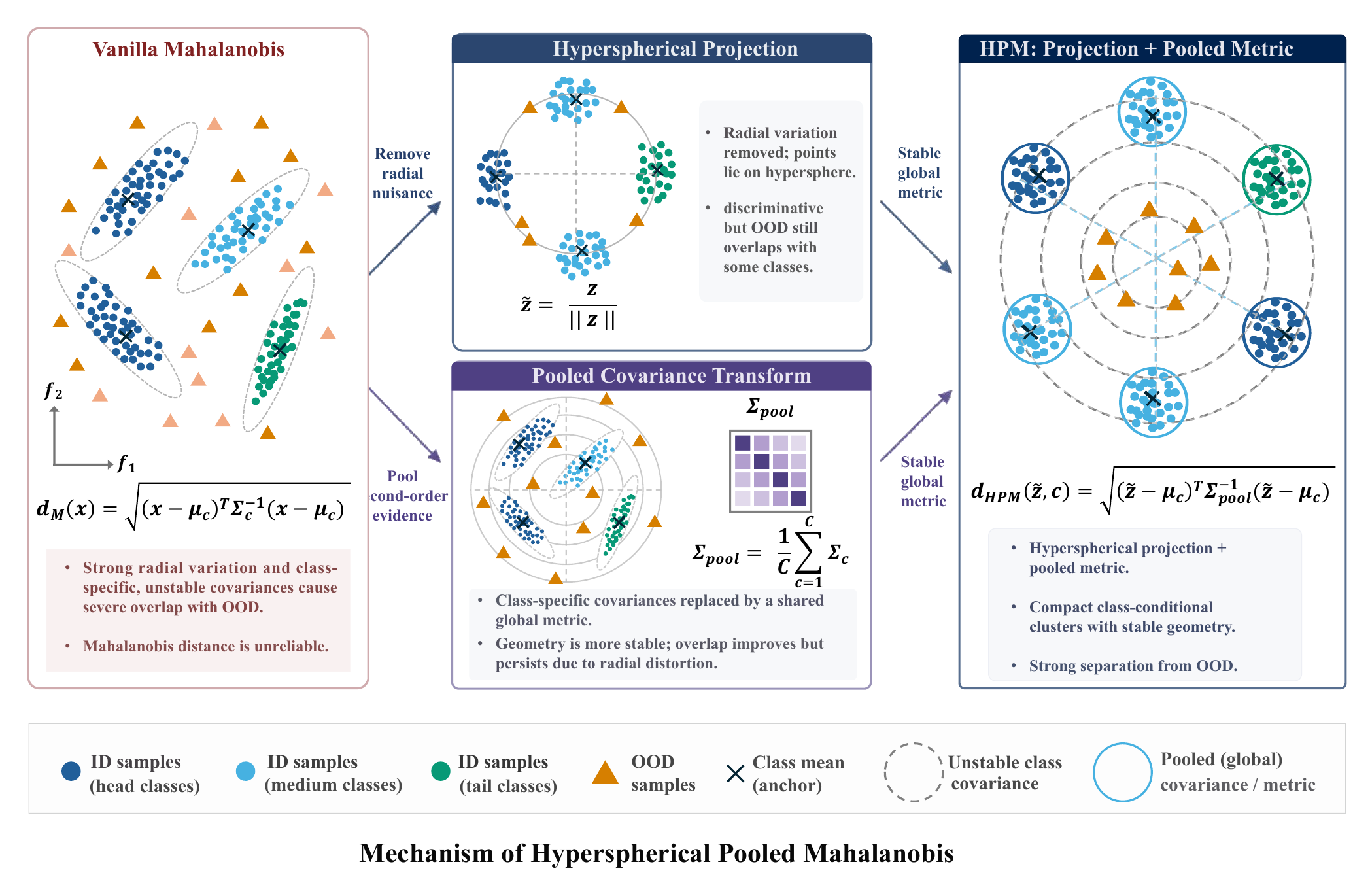}
    \caption{\textbf{\HPM geometry repair.} Class-conditional Mahalanobis is unstable when tail classes lack residual support (top path). \HPM first projects features to the hypersphere to remove radial bias correlated with imbalance, then uses pooled residual covariance to replace tail-specific spectra with shared, ridge-regularized geometry while preserving class mean anchors.}
    \label{fig:metric-repair}
\end{figure}

We propose \emph{Hyperspherical Pooled Mahalanobis} (\HPM), a minimal post-hoc detector that repairs this geometry. It maps features onto the unit hypersphere to remove frequency-coupled radius, then replaces unsupported class-specific covariance with a pooled ridge covariance. Class means remain semantic anchors, while second-order geometry is estimated from shared within-class residuals. Figure~\ref{fig:metric-repair} summarizes the mechanism. \HPM changes only the detector: it adds no outlier head, auxiliary OOD supervision, model retraining, or specialized training objective.

\HPM also complements classifier-score diagnostics. Energy and maximum softmax probability (MSP) depend only on classifier outputs, so they cannot respond to feature variation that leaves logits unchanged. This classifier-null blind spot is not our main claim, but it explains why post-hoc LT-OOD evaluation should not rely on classifier scores alone. A repaired feature-space metric can expose evidence that raw Mahalanobis distorts and classifier scores cannot observe.

Our evaluations on CIFAR-10-LT, CIFAR-100-LT, and ImageNet-100-LT use representations from recent LT-OOD and long-tail methods including PATT~\citep{he2025patt}, PASCL~\citep{wang2022pascl}, AdaptOD~\citep{miao2024adaptod}, COCL~\citep{miao2024cocl}, EAT~\citep{wei2024eat}, and DARL~\citep{zhang2025darl}. On PC-ERM, \HPM raises raw Mahalanobis AUROC from 46.49 to 85.67 on CIFAR-10-LT and from 50.40 to 78.35 on CIFAR-100-LT. PC-ERM+\HPM also achieves the highest CIFAR-100-LT LES (3.08), retaining roughly 95\% of the best CIFAR-100-LT AUROC observed among the compared post-hoc scores at substantially lower training-time cost. ImageNet-100-LT acts as a boundary analysis of how far stabilized Mahalanobis geometry can repair the raw feature-space detector.

Our contributions are:
\begin{itemize}
    \item We revisit whether increasingly complex training is necessary for LT-OOD detection, and use LES to evaluate OOD performance together with training-time cost.
    \item We identify two geometric reasons why raw Mahalanobis fails under long-tailed representations: frequency-coupled feature radius and poorly supported tail-class covariance estimation.
    \item We propose \HPM, a post-hoc Mahalanobis repair based on hyperspherical projection and pooled ridge covariance, showing that simple PC-ERM models can achieve strong and efficient OOD detection without auxiliary OOD supervision or retraining.
\end{itemize}

\section{Related Work}

\paragraph{Post-hoc OOD detection and Mahalanobis geometry.}
Classical post-hoc OOD detectors score softmax confidence~\citep{hendrycks2017baseline}, use input preprocessing or temperature scaling~\citep{liang2018odin}, or measure feature-space density with Mahalanobis distance~\citep{lee2018mahalanobis}. Energy scoring~\citep{liu2020energy} has become a strong classifier-score baseline, while later detectors exploit feature clipping, nearest neighbors, residual subspaces, or feature compactness~\citep{sun2021react,sun2022knn,wang2022vim,sehwag2021ssd,ming2023cider}. Our work is closest to Mahalanobis-style detection and recent feature-normalized variants~\citep{muller2025mahalanobispp}, but studies a different failure mode: under long-tailed training, raw feature radius is frequency-coupled and tail classes cannot support reliable class-specific covariance.

\paragraph{Long-tailed recognition and LT-OOD detection.}
Long-tailed recognition has been addressed through re-weighting, re-sampling, margin adjustment, logit adjustment, contrastive learning, and decoupled training~\citep{cao2019ldam,kang2020decoupling,menon2021longtail,ren2020balanced,zhu2022balanced,li2022targeted}. LT-OOD detection adds the open-set requirement that rare ID samples remain accepted while unknown samples are rejected. Recent methods improve this setting with contrastive tail-OOD separation, abstention or calibrated outlier classes, adaptive outlier distributions, gradient-conflict mitigation, and semantic tail prioritization~\citep{wang2022pascl,wei2024eat,miao2024cocl,miao2024adaptod,zhang2025darl,he2025patt}. These approaches mainly change training. We ask how far a stable post-hoc geometry can take simple long-tail training before adding more specialized objectives.

\paragraph{Auxiliary OOD supervision and representation geometry.}
Outlier exposure can greatly improve OOD detection when suitable auxiliary data are available~\citep{hendrycks2019oe}, but it also couples OOD rejection with the learned ID geometry. This coupling is especially delicate in long-tailed learning, where head and tail classes receive very different statistical support. Our analysis connects LT-OOD behavior to classifier-null scatter, feature radius, covariance spectra, and effective rank, complementing studies of feature geometry and neural collapse~\citep{papyan2020neuralcollapse,kornblith2019better}.

\section{Preliminaries}

\paragraph{LT-OOD setting.}
Let $\mathcal{D}_{\ID}=\{(x_i,y_i)\}$ be a long-tailed training set with $K$ ID classes and class counts $\{n_c\}_{c=1}^K$. At test time, a detector receives a query $x$ and decides whether it comes from the ID distribution or from an unseen OOD distribution. We use the convention that larger OOD scores indicate samples that are more likely OOD.

\paragraph{Energy score.}
Given logits $f(x)\in\R^K$ and temperature $T>0$, the energy score is
\begin{equation}
    \Energy(x)=-T\log\sum_{k=1}^{K}\exp(f_k(x)/T).
\end{equation}

\paragraph{Classifier row space and null space.}
Let $h(x)\in\R^d$ be the penultimate feature and let the linear classifier be
$f(h)=Wh+b$, where $W=[w_1^\top;\ldots;w_K^\top]\in\R^{K\times d}$.
The classifier row space is $\mathcal{R}(W^\top)$, and its orthogonal complement is the classifier-null space $\mathcal{N}(W)=\{v:Wv=0\}$. We use the orthogonal projectors
\begin{equation}
    P_{\row}=W^\top(WW^\top)^\dagger W,
    \qquad
    P_{\nullsp}=I-P_{\row}.
\end{equation}
For any feature $h=P_{\row}h+P_{\nullsp}h$, we have $WP_{\nullsp}h=0$. Therefore,
\begin{equation}
    f(h)=Wh+b=WP_{\row}h+b=f(P_{\row}h),
    \qquad
    \Energy(h)=\Energy(P_{\row}h).
\end{equation}
Thus classifier scores, including Energy, are invariant to any perturbation that lies in the classifier-null component $P_{\nullsp}h$.

\paragraph{Classifier-null absolute scatter.}
To quantify the amount of feature variation that Energy cannot observe, we measure the within-class scatter that lies in the classifier-null subspace. For class $c$, let
\begin{equation}
    \widehat{\mu}_c=\frac{1}{n_c}\sum_{i:y_i=c} h_i,
    \qquad
    \widehat{\Sigma}_c=\frac{1}{n_c-1}\sum_{i:y_i=c}(h_i-\widehat{\mu}_c)(h_i-\widehat{\mu}_c)^\top .
\end{equation}
We define the absolute classifier-null scatter as
\begin{equation}
    A_c^{\nullsp}
    =\mathrm{Tr}\!\left(P_{\nullsp}\widehat{\Sigma}_cP_{\nullsp}\right)
    =\frac{1}{n_c-1}\sum_{i:y_i=c}
    \left\|P_{\nullsp}(h_i-\widehat{\mu}_c)\right\|_2^2 .
    \label{eq:null-absolute-scatter}
\end{equation}
This quantity is an absolute trace rather than a fraction. It measures how much class-conditional variation is hidden from classifier scores, not just what proportion of the total scatter lies in the null space. For a group of classes $\mathcal{G}$, such as head or tail classes, we report
\begin{equation}
    A_{\mathcal{G}}^{\nullsp}
    =\frac{1}{|\mathcal{G}|}\sum_{c\in\mathcal{G}} A_c^{\nullsp}.
\end{equation}

\section{Diagnosing and Repairing Long-Tailed Mahalanobis Geometry}

\subsection{Energy has a classifier-null blind spot}

Let $h(x)\in\R^d$, logits $f(x)=Wh(x)+b$, and
\begin{equation}
    \Energy(x)=-T\log\sum_{c=1}^{K}\exp(f_c(x)/T).
\end{equation}
For any $\delta\in\nullsp(W)$, $W(h+\delta)+b=Wh+b$. Thus Energy observes only classifier-visible coordinates. It can be strong when OOD evidence is classifier-aligned, but it is blind to classifier-null feature variation.

\subsection{Raw Mahalanobis mixes feature radius with geometry}

For raw features $h=ru$, $r=\|h\|_2$, and $\|u\|_2=1$,
\begin{equation}
    \mathrm{Cov}(h)=\E[r^2uu^\top]-\E[ru]\E[ru]^\top.
\end{equation}
When $r$ is coupled with class frequency, raw Mahalanobis mixes semantic displacement with a long-tail nuisance coordinate. We score in hyperspherical coordinates $z=h/\|h\|_2$ to remove this radius term before computing distances.

\subsection{Tail classes cannot support class-specific covariance}

A class-specific covariance in $d$ dimensions has
\begin{equation}
    \mathrm{rank}(\widehat{\Sigma}_c)\le n_c-1.
\end{equation}
For tail classes, $n_c\ll d$ is common, so full class-specific precision estimates have little support. HPM keeps class means specific but pools second-order geometry.

\subsection{Why hyperspherical projection and pooled covariance repair the metric}
\label{sec:hpm-theory}

\begin{lemma}[Pooled covariance expands supported directions]
\label{lem:pooled-rank}
Assume $n_c\ge2$ for each class. Let
\begin{equation}
    \widehat{\mu}_c^z=\frac{1}{n_c}\sum_{i:y_i=c}z_i,\qquad
    \widehat{\Sigma}_c^z=\frac{1}{n_c-1}\sum_{i:y_i=c}(z_i-\widehat{\mu}_c^z)(z_i-\widehat{\mu}_c^z)^\top ,
\end{equation}
let $S_c=\mathrm{span}\{z_i-\widehat{\mu}_c^z:y_i=c\}$, and let
\begin{equation}
    \widehat{\Sigma}_{\mathrm{pool}}=\sum_{c=1}^{K}\alpha_c\widehat{\Sigma}_c^z,\qquad
    \alpha_c>0,\quad \sum_c\alpha_c=1 .
\end{equation}
Then
\begin{equation}
    \mathrm{range}(\widehat{\Sigma}_{\mathrm{pool}})
    =
    \mathrm{span}\Big(\bigcup_{c=1}^{K}S_c\Big).
\end{equation}
With ridge regularization, $\widehat{\Sigma}_{\mathrm{pool}}+\lambda I\succ 0$ for any $\lambda>0$.
\end{lemma}

\begin{theorem}[HPM as stabilized Mahalanobis geometry]
\label{thm:hpm-geometry}
Assume $N>K$ and $n_c\ge2$ for each class. Let $z=h/\|h\|_2$ and define
\begin{equation}
    \widehat{\Sigma}_{\HPM}
    =
    \frac{1}{N-K}\sum_{c=1}^{K}\sum_{i:y_i=c}(z_i-\widehat{\mu}_c^z)(z_i-\widehat{\mu}_c^z)^\top+\lambda I,
    \qquad \lambda>0 .
\end{equation}
Then \HPM uses normalized class means as semantic anchors, removes raw feature radius before covariance estimation, and replaces tail-specific covariance with pooled residual geometry. Lemma~\ref{lem:pooled-rank} characterizes the covariance support before ridge regularization; after adding $\lambda I$, the precision is full-rank, with residual-supported directions shaped by the pooled covariance and orthogonal directions controlled only by the isotropic ridge.
\end{theorem}

The result is a mechanism statement rather than a universal dominance claim.

\subsection{The repaired detector}

Given the normalized class means and pooled ridge covariance above, the OOD score is
\begin{equation}
    s_{\HPM}(x)=\min_c \left(\frac{h(x)}{\|h(x)\|_2}-\widehat{\mu}_c^z\right)^\top\widehat{\Sigma}_{\HPM}^{-1}\left(\frac{h(x)}{\|h(x)\|_2}-\widehat{\mu}_c^z\right).
\end{equation}

\begin{algorithm}[t]
\caption{\HPM post-hoc OOD scoring. The algorithm normalizes frozen ID and query features, estimates class anchors with a pooled ridge covariance, and returns a Mahalanobis score where larger values indicate more OOD-like samples.}
\label{alg:hpm}
\small
\begin{algorithmic}[1]
    \Require Frozen encoder $h(\cdot)$; ID training set $\{(x_i,y_i)\}_{i=1}^{N}$ with labels $y_i\in\{1,\ldots,K\}$; query sample $x$; ridge penalty $\lambda>0$.
    \Ensure Post-hoc OOD score $s_{\HPM}(x)$ (larger implies more $\OOD$-like).
    \State Collect training features $\{h_i\}_{i=1}^{N}$ with $h_i\gets h(x_i)$ and the query feature $h\gets h(x)$.
    \State Project onto the hypersphere $z_i\gets h_i/\|h_i\|_2$ and $z\gets h/\|h\|_2$.
    \State Estimate class anchors $\{\widehat{\mu}_c^z\}_{c=1}^{K}$ and pooled ridge covariance $\widehat{\Sigma}_{\HPM}$ exactly as in Theorem~\ref{thm:hpm-geometry}.
    \State \Return $s_{\HPM}(x)=\min_{c\in[K]} \left(z-\widehat{\mu}_c^z\right)^\top\widehat{\Sigma}_{\HPM}^{-1}\left(z-\widehat{\mu}_c^z\right)$.
\end{algorithmic}
\end{algorithm}

\section{Experiments}

\subsection{Setup}

\paragraph{Datasets and models.}
We evaluate on CIFAR-10-LT, CIFAR-100-LT~\citep{krizhevsky2009learning}, and ImageNet-100-LT~\citep{liu2019imagenetlt,russakovsky2015imagenet}. The CIFAR-LT OOD suite follows common OOD benchmarks~\citep{yang2022openood}, covering digit~\citep{netzer2011reading}, texture~\citep{cimpoi2014describing}, scene~\citep{yu2015lsun}, and object-level shifts; ImageNet-100-LT uses held-out ImageNet classes as near-OOD samples. CIFAR baselines include PATT~\citep{he2025patt}, PASCL~\citep{wang2022pascl}, DARL~\citep{zhang2025darl}, AdaptOD~\citep{miao2024adaptod}, COCL~\citep{miao2024cocl}, EAT~\citep{wei2024eat}, and PC-ERM. For methods originally trained with outlier exposure (OE), we report both the original checkpoint and the matched no-OE variant when available. PC-ERM uses the same backbone family and 200-epoch CIFAR budget, but only standard cross-entropy plus long-tail log-prior adjustment~\citep{menon2021longtail}; it has no auxiliary OOD data, outlier head, contrastive OOD objective, or detector-specific training loss.

\paragraph{Scores and unified post-hoc protocol.}
We compare Energy, maximum softmax probability (MSP), standard Mahalanobis, raw pooled Mahalanobis (\PMD), hyperspherical class-specific Mahalanobis (\SMD), and \HPM under a common post-hoc OOD averaging protocol. For CIFAR-LT, we average over the OOD sets listed above. OOD detection is reported with area under the receiver operating characteristic (AUROC) and false positive rate at 95\% true positive rate (FPR95); higher AUROC and lower FPR95 are better. ACC denotes closed-set ID classification accuracy. We also report LES,
\begin{equation}
    \mathrm{LES}=\lg\left(\frac{\mathrm{BestAUROC}}{\mathrm{Cost}}\right).
\end{equation}
where BestAUROC is the strongest AUROC in the row and Cost is the model training time.

\paragraph{Mahalanobis variants.}
Let $\mu_c$ be the feature mean of class $c$. A Mahalanobis detector scores a query by
\begin{equation}
    s(x)=\min_{c} (z(x)-\mu_c)^\top \Sigma_c^{-1}(z(x)-\mu_c),
\end{equation}
where $z(x)$ may be the raw feature $h(x)$ or its normalized version $h(x)/\|h(x)\|_2$, and $\Sigma_c$ may be class-specific or pooled across classes. Table~\ref{tab:mahalanobis-variants} summarizes the four post-hoc variants. Our method is \HPM, which combines hyperspherical projection with pooled covariance.

\subsection{Classifier-null blind spot}
\label{sec:nullspace-experiment}

\begin{figure}[t]
\centering
\begin{minipage}[t]{0.58\textwidth}
\centering
\vspace{0pt}
\subcaptionbox{CIFAR-10-LT\label{fig:nullspace-c10}}{\includegraphics[width=0.49\linewidth,trim=0 0 327 0,clip]{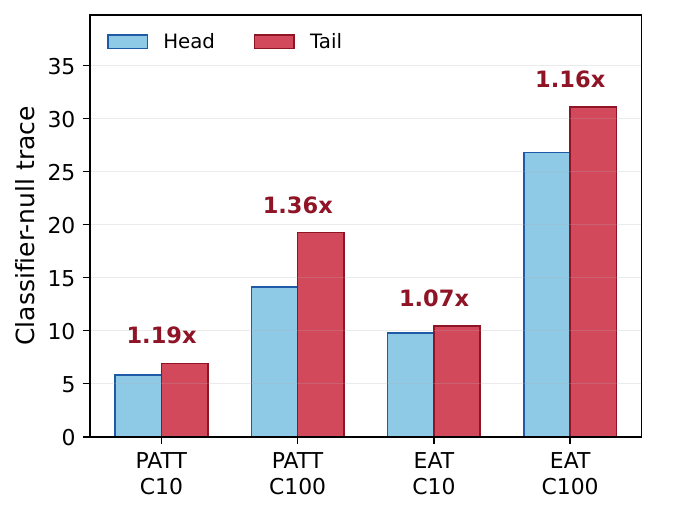}}\hfill
\subcaptionbox{CIFAR-100-LT\label{fig:nullspace-c100}}{\includegraphics[width=0.49\linewidth,trim=327 0 0 0,clip]{figures/fig_nullspace_head_tail_bar.pdf}}
\caption{Classifier-null scatter in CIFAR-LT.}
\label{fig:nullspace-bar}
\end{minipage}\hfill
\begin{minipage}[t]{0.38\textwidth}
\centering
\vspace{0pt}
\captionof{table}{\textbf{Mahalanobis variants.} Feature normalization and covariance estimation choices for the four post-hoc distance scores.}
\label{tab:mahalanobis-variants}
\footnotesize
\setlength{\tabcolsep}{2.6pt}
\begin{tabular}{@{}lll@{}}
\toprule
Name & Feature & Covariance \\
\midrule
\VMD & raw & class-specific \\
\PMD & raw & pooled \\
\SMD & hyperspherical & class-specific \\
\HPM & hyperspherical & pooled \\
\bottomrule
\end{tabular}
\end{minipage}
\end{figure}

The following diagnostics explain why the same frozen representation can lead to different conclusions under classifier-score and feature-space detectors. Energy is computed only from classifier logits, so it cannot respond to variation orthogonal to the classifier row space. Figure~\ref{fig:nullspace-bar} measures this blind spot in CIFAR-LT checkpoints: tail classes often carry larger classifier-null scatter than head classes, and the null-space component varies systematically with class count.

This result does not say that classifier scores are weak. Instead, it identifies a channel they cannot read. When OOD evidence is aligned with the classifier, Energy or MSP can be strong; when evidence lies in classifier-null variation, a feature-space detector has access to information that classifier outputs discard.

\subsection{Feature-radius coupling}
\label{sec:radius-pollution}

\begin{figure}[t]
    \centering
    \includegraphics[width=0.96\linewidth,trim=0 18 0 0,clip]{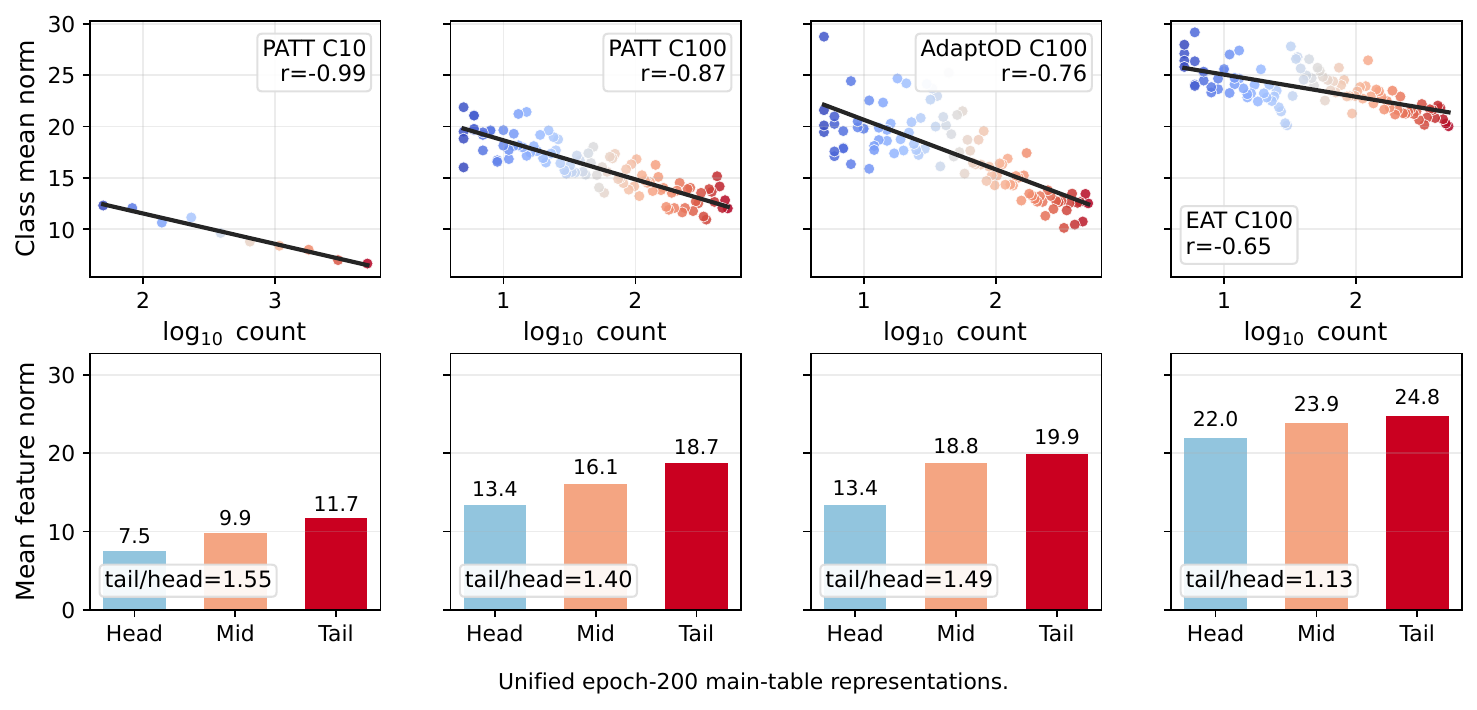}
    \caption{\textbf{Feature radius tracks class frequency.} Per-class backbone-feature norm versus training frequency across representative CIFAR-LT checkpoints used in the main tables. Larger norms for lower-frequency classes indicate that raw covariance is fit in coordinates where radius is coupled to imbalance, motivating the hyperspherical projection used before scoring.}
    \label{fig:norm}
\end{figure}

Figure~\ref{fig:norm} shows the radius-frequency coupling behind hyperspherical projection. Across representative checkpoints, raw feature norms increase from head to tail groups; for example, the tail/head ratio ranges from about $1.13$ to $1.49$ in the plotted cases. Thus raw Mahalanobis can treat class-frequency-induced radius variation as part of the distance geometry.

This explains why feature-space detection still needs repair. A larger radius for rare classes is not necessarily semantic OOD evidence; it can be a training artifact of long-tailed supervision. Hyperspherical projection removes this radial degree of freedom before covariance estimation, forcing the score to compare angular residual structure rather than raw norm scale.

\subsection{Covariance distortion}
\label{sec:covariance-distortion}

\begin{figure}[t]
    \centering
    \includegraphics[width=0.98\linewidth]{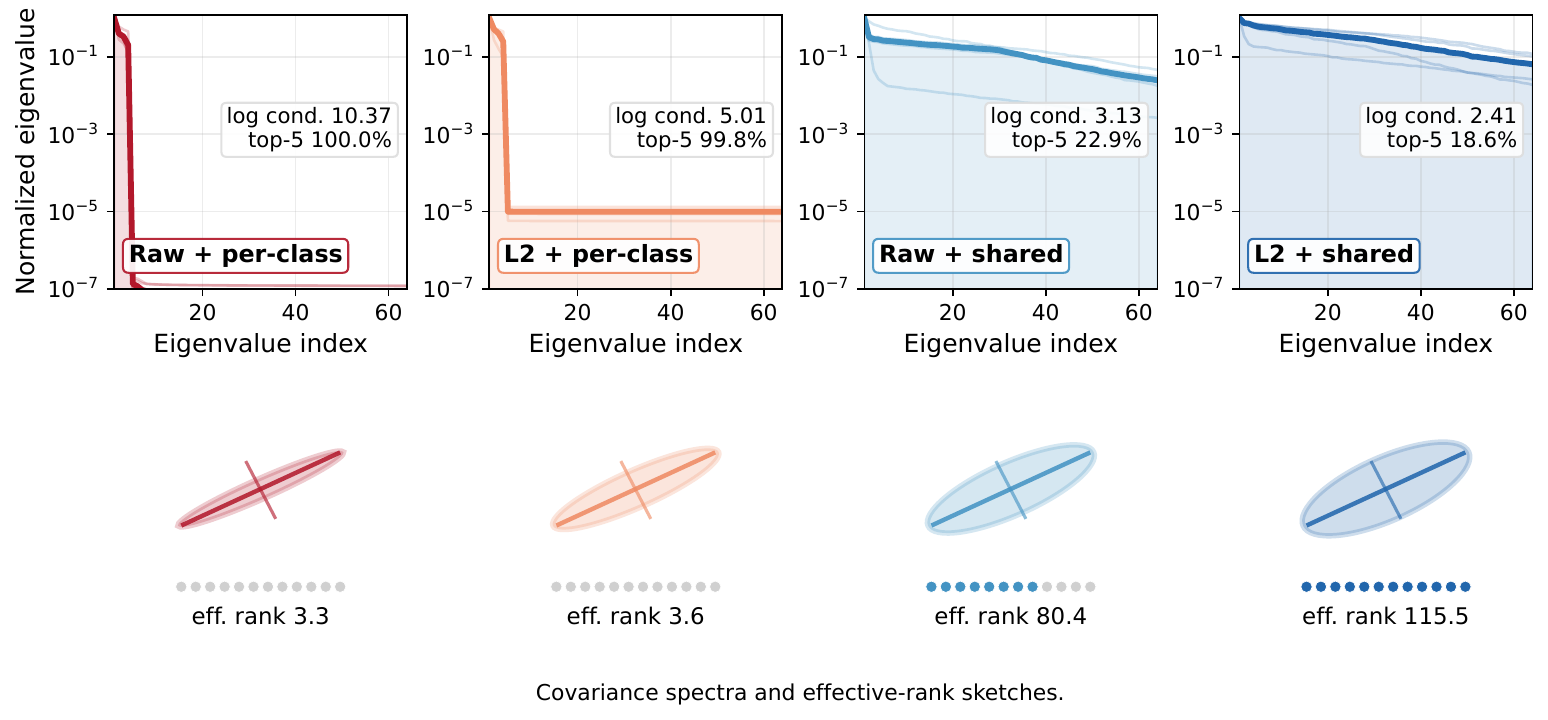}
    \caption{\textbf{Pooled hyperspherical covariance is better supported.} Four Mahalanobis variants on CIFAR-100-LT feature banks. Top: eigenvalues of the quadratic form, where sharp drops indicate dependence on a few dominant axes. Bottom: elliptical level sets with effective-rank and conditioning summaries; higher effective rank and lower condition number indicate better-supported geometry.}
    \label{fig:covariance}
\end{figure}

Figure~\ref{fig:covariance} shows the second failure mode: raw per-class covariance is dominated by a few unstable directions. Across PATT, PC-ERM, DARL, and PASCL on CIFAR-100-LT, the median effective rank rises from about $3.3$ for raw per-class covariance to about $115.5$ for pooled hyperspherical covariance, while the median log-condition number drops from about $10.4$ to about $2.4$.

The comparison isolates the role of the two design choices in Table~\ref{tab:mahalanobis-variants}. Hyperspherical projection removes the radial nuisance identified in Figure~\ref{fig:norm}; pooling then replaces tail-specific covariance estimates with residual directions shared across classes. This is consistent with Lemma~\ref{lem:pooled-rank}: pooling does not create new features, but it aggregates complementary residual support; empirically, this produces a better-conditioned metric in Figure~\ref{fig:covariance}.

\paragraph{Outlier exposure as a side analysis.}
We treat OE as a side analysis rather than as the main target. Controlled no-OE/light-OE comparisons show that OE changes ID compactness, classifier-null variation, and radius-covariance coupling. Thus an OOD objective can improve a task-specific score without necessarily producing cleaner post-hoc geometry; this motivates reporting detector behavior on the same frozen checkpoints in the main tables.

\subsection{CIFAR-LT results}
\label{sec:cifar-results}

Tables~\ref{tab:cifar10-unified-posthoc} and~\ref{tab:cifar100-unified-posthoc} compare six post-hoc scores on the same frozen CIFAR-LT representations. This tests when \HPM repairs raw Mahalanobis scoring and when the repaired feature-space detector remains competitive with classifier scores.

\begin{table}[t]
\centering
\caption{\textbf{CIFAR-10-LT post-hoc OOD results.} Entries are averaged AUROC\,/\,FPR95; detector columns mark row-wise best/second-best values, ACC/LES are column-wise, and only top-three \HPM cells are shaded.}
\label{tab:cifar10-unified-posthoc}
\small
\resizebox{\textwidth}{!}{%
\begin{tabular}{lcccccccc}
\toprule
Model & Energy & MSP & \VMD & \PMD & \SMD & \HPM & ACC & LES$\uparrow$ \\
\midrule
\multicolumn{9}{l}{\emph{Models trained with auxiliary outlier exposure.}}\\
PATT & 81.78/51.29 & \textbf{86.25/46.39} & 46.61/87.99 & 56.80/76.80 & 82.76/47.57 & \secondcell{85.20/44.28} & 79.86 & 1.96 \\
AdaptOD & 66.72/72.49 & 67.36/74.30 & 55.07/84.78 & 58.98/86.13 & \underline{72.97/68.10} & \bestcell{77.75/62.88} & 57.57 & 2.84 \\
COCL & \textbf{89.75/42.17} & 88.21/46.08 & 62.31/65.99 & 89.04/42.44 & 87.75/43.32 & \secondcell{89.20/41.93} & 80.08 & 2.84 \\
EAT & \underline{78.29/59.07} & \textbf{79.57/60.21} & 61.31/83.54 & 60.40/83.07 & 74.14/66.20 & 70.22/71.57 & 79.50 & 1.83 \\
\multicolumn{9}{l}{\emph{Models trained without auxiliary outlier exposure.}}\\
PATT & \textbf{90.17/36.17} & \underline{87.47/40.06} & 58.28/76.70 & 55.30/78.17 & 85.44/43.34 & \thirdcell{86.78/41.05} & 83.01 & 1.97 \\
PASCL & \underline{87.08/42.48} & 83.53/46.15 & 66.01/76.59 & 66.09/76.80 & 82.38/49.27 & \bestcell{88.89/40.29} & \underline{85.78} & 2.40 \\
DARL & 77.20/66.26 & \underline{83.78/54.22} & 34.28/90.57 & 55.20/86.60 & 76.39/53.49 & \bestcell{88.72/38.15} & \textbf{87.71} & 2.48 \\
AdaptOD & 73.87/64.66 & 69.48/67.98 & 66.98/77.81 & 57.79/81.70 & \underline{78.52/61.51} & \bestcell{79.36/59.54} & 66.58 & \underline{2.87} \\
COCL & \textbf{76.38/59.63} & \underline{74.15/61.11} & 72.04/65.71 & 59.36/76.31 & 72.86/60.71 & 70.58/67.03 & 76.81 & 2.79 \\
EAT & 65.39/80.19 & 64.34/79.98 & \underline{66.50/82.82} & 55.82/82.95 & \textbf{68.16/74.35} & \thirdcell{65.58/78.49} & 73.40 & 1.77 \\
PC-ERM & \underline{85.06/50.50} & 81.32/50.56 & 46.49/92.15 & 45.39/90.61 & 77.65/53.60 & \bestcell{85.67/46.38} & 84.09 & \textbf{3.05} \\
\bottomrule
\end{tabular}%
}
\end{table}

\begin{table}[t]
\centering
\caption{\textbf{CIFAR-100-LT post-hoc OOD results.} Same protocol and notation as Table~\ref{tab:cifar10-unified-posthoc}.}
\label{tab:cifar100-unified-posthoc}
\small
\resizebox{\textwidth}{!}{%
\begin{tabular}{lcccccccc}
\toprule
Model & Energy & MSP & \VMD & \PMD & \SMD & \HPM & ACC & LES$\uparrow$ \\
\midrule
\multicolumn{9}{l}{\emph{Models trained with auxiliary outlier exposure.}}\\
PATT & 73.86/57.89 & \underline{78.73/53.96} & 65.30/70.76 & 44.13/90.77 & 75.72/58.08 & \bestcell{81.41/46.81} & 48.22 & 2.03 \\
AdaptOD & \textbf{67.22/73.76} & \underline{59.85/84.38} & 43.84/92.25 & 44.55/91.80 & 57.24/83.94 & 51.42/89.33 & 32.98 & 2.80 \\
COCL & \textbf{73.46/68.85} & 71.03/72.49 & 44.98/93.28 & 65.93/75.57 & 68.91/71.93 & \secondcell{71.40/69.95} & 45.10 & 2.77 \\
EAT & \textbf{68.04/75.26} & 62.97/82.20 & 57.19/89.57 & 58.03/85.57 & 62.92/81.10 & \secondcell{63.39/83.81} & 43.70 & 1.80 \\
\multicolumn{9}{l}{\emph{Models trained without auxiliary outlier exposure.}}\\
PATT & \underline{80.29/52.72} & 79.75/55.40 & 70.79/71.07 & 41.33/92.46 & 76.13/60.38 & \bestcell{82.61/45.56} & 51.33 & 2.03 \\
PASCL & \underline{76.20/64.42} & 70.74/72.87 & 55.71/88.75 & 54.16/89.30 & 73.64/64.55 & \bestcell{79.30/58.46} & 50.88 & 2.43 \\
DARL & 62.17/78.50 & 68.54/74.88 & 53.95/90.93 & 47.33/93.39 & \underline{69.64/71.15} & \bestcell{79.31/59.35} & \textbf{56.63} & 2.25 \\
AdaptOD & \underline{70.96/71.68} & 65.20/78.56 & 61.27/82.52 & 40.59/95.10 & 69.02/74.84 & \bestcell{72.82/69.86} & 35.86 & \underline{2.86} \\
COCL & \underline{70.04/72.35} & 66.22/77.32 & 42.21/93.50 & 52.29/84.81 & 69.36/74.41 & \bestcell{76.86/66.57} & 45.99 & 2.82 \\
EAT & 63.60/81.13 & 61.76/83.14 & 60.77/85.63 & 52.55/89.40 & \underline{65.50/76.72} & \bestcell{70.09/75.21} & 41.39 & 1.82 \\
PC-ERM & \underline{73.44/67.64} & 67.96/73.05 & 50.40/91.26 & 54.81/90.77 & 71.95/66.46 & \bestcell{78.35/56.92} & \underline{51.91} & \textbf{3.08} \\
\bottomrule
\end{tabular}%
}
\end{table}

\paragraph{Result analysis.}
\HPM substantially improves raw Mahalanobis in most rows and is the best detector for all no-OE CIFAR-100-LT rows. Energy and MSP remain stronger in some CIFAR-10-LT or OE-trained cases, showing that classifier evidence and repaired feature geometry are complementary. For PC-ERM, \HPM is best on both CIFAR datasets, raising raw Mahalanobis from 46.49 to 85.67 AUROC on CIFAR-10-LT and from 50.40 to 78.35 AUROC on CIFAR-100-LT, while also giving the highest LES. Figure~\ref{fig:cifar100-les} shows that this simple classifier retains roughly 95\% of the best CIFAR-100-LT AUROC observed among the compared post-hoc scores with much lower training-time cost.

\subsection{ImageNet-100-LT near-OOD analysis}
\label{sec:imagenet-results}

Table~\ref{tab:imagenet100-boundary} compares raw \VMD with \HPM on ImageNet-100-LT. Raw Mahalanobis remains consistently weak, while \HPM substantially repairs feature-space scoring. This makes ImageNet-100-LT a boundary case for how much stabilization alone can recover from the raw Mahalanobis detector.

\begin{center}
\centering
\begin{minipage}[t]{0.54\textwidth}
\centering
\vspace{0pt}
\includegraphics[width=\linewidth]{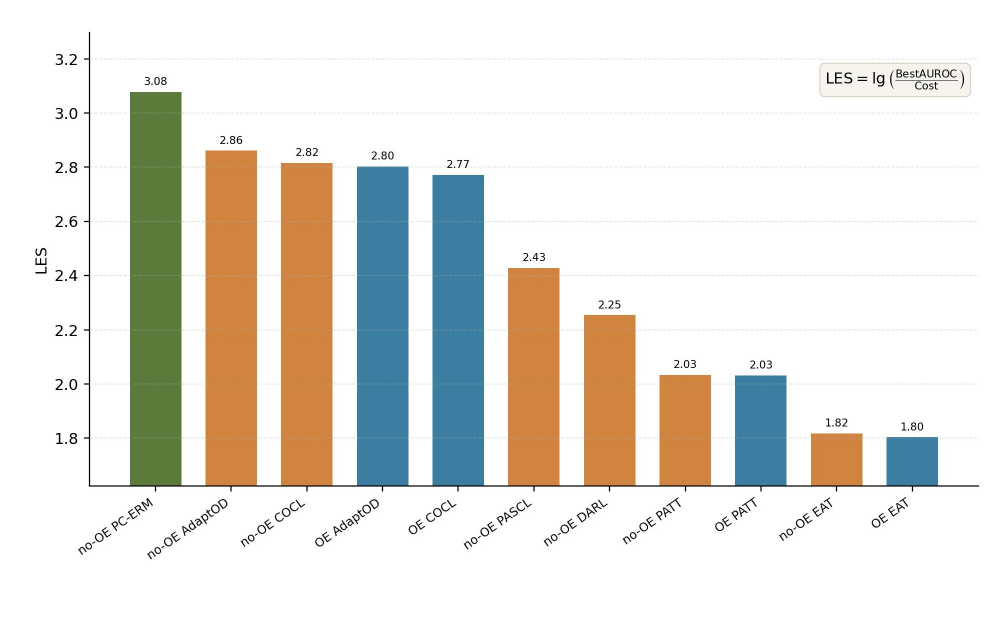}
\captionof{figure}{\textbf{CIFAR-100-LT efficiency.} LES compares best AUROC with training time; higher is better.}
\label{fig:cifar100-les}
\end{minipage}\hfill
\begin{minipage}[t]{0.42\textwidth}
\centering
\vspace{0pt}
\captionsetup{justification=raggedright,singlelinecheck=false}
\captionof{table}{\textbf{ImageNet-100-LT near-OOD.} Main AUROC/FPR95 comparison of \VMD and \HPM.}
\label{tab:imagenet100-boundary}
\scriptsize
\setlength{\tabcolsep}{1.8pt}
\resizebox{\linewidth}{!}{%
\begin{tabular}{lcccc}
\toprule
Model & \VMD & \HPM & ACC & LES$\uparrow$ \\
\midrule
\multicolumn{5}{l}{\emph{Models trained with auxiliary outlier exposure.}}\\
PATT & \underline{46.73/93.03} & \bestcell{81.04/54.33} & \underline{59.70} & 0.97 \\
COCL & \underline{41.74/90.34} & \bestcell{81.45/59.35} & 42.45 & 1.96 \\
EAT & \underline{57.36/91.26} & \bestcell{80.18/63.43} & 39.35 & 1.67 \\
\multicolumn{5}{l}{\emph{Models trained without auxiliary outlier exposure.}}\\
PATT & \underline{40.99/95.65} & \bestcell{80.04/54.97} & 55.60 & 0.95 \\
COCL & \underline{35.26/95.96} & \bestcell{77.70/60.22} & 41.20 & \underline{1.97} \\
EAT & \underline{50.47/95.29} & \bestcell{74.92/66.76} & 38.70 & 1.64 \\
PASCL & \underline{43.17/95.96} & \bestcell{88.26/37.88} & \textbf{66.30} & 1.04 \\
PC-ERM & \underline{52.85/89.42} & \bestcell{63.47/84.14} & 46.40 & \textbf{2.42} \\
\bottomrule
\end{tabular}%
}
\end{minipage}
\end{center}

\section{Discussion and Limitations}
\label{sec:discussion}

Energy is effective when OOD separation is visible to the classifier, while \HPM reads evidence that becomes usable after stabilizing feature geometry. LT-OOD performance should therefore separate the representation, the detector, and the cost paid to train the model.

PC-ERM suggests that gains from specialized LT-OOD training should be re-evaluated against simple training plus repaired post-hoc geometry. Auxiliary OOD data, extra heads, contrastive losses, energy regularization, and gradient-conflict mitigation can help, but in several CIFAR comparisons the gap narrows once detector geometry is stabilized. With only cross-entropy, long-tail prior calibration, and \HPM, PC-ERM remains competitive on CIFAR-LT and obtains the best LES in both condensed CIFAR comparisons. On CIFAR-100-LT, it retains about 95\% of the best AUROC observed among the compared post-hoc scores while achieving the highest LES.
\HPM is not a universal substitute for training-time objectives. ImageNet-100-LT remains a boundary case for stabilized Mahalanobis geometry: \HPM repairs raw \VMD substantially, but this repair alone does not settle all near-OOD settings. Our results instead argue that new objectives should justify their added complexity against simple baselines such as PC-ERM, stable post-hoc geometry, and efficiency metrics such as LES.

\section{Conclusion}

We revisited whether strong LT-OOD detection requires increasingly complex training. Raw class-specific Mahalanobis distance is distorted by feature-radius nuisance and unsupported tail covariance, so a simple classifier can be underrated when evaluated with an unstable feature-space metric. \HPM addresses this geometry without auxiliary OOD data, retraining, or a new objective. Paired with simple PC-ERM, this repair yields competitive OOD performance and high LES, showing that new LT-OOD objectives should justify their added cost against stable post-hoc geometry. Code is available at \url{https://anonymous.4open.science/r/Neurips-26-DE01/}.

\bibliographystyle{plainnat}
\bibliography{references}

\end{document}